\documentclass[lettersize,journal]{IEEEtran}
\usepackage{amsmath,amsfonts}
\usepackage{algorithmic}
\usepackage{algorithm}
\usepackage{array}
\usepackage{svg}
\usepackage[caption=false,font=normalsize,labelfont=sf,textfont=sf]{subfig}
\usepackage{textcomp}
\usepackage{stfloats}
\usepackage{url}
\usepackage{verbatim}
\usepackage{graphicx}
\usepackage{cite}
\usepackage{multirow}
\usepackage[skip=4pt]{caption}
\begin{document}
\title{Uncertainty Quantification for EO Regression Tasks: Building Height, Tree Canopy Height and Above-ground Biomass Estimation}

\author{Ritu~Yadav, Andrea~Nascetti, and Yifang~Ban%
\thanks{Ritu Yadav, Andrea Nascetti and Yifang Ban is with Division of Geoinformatics, KTH Royal Institute of Technology, Sweden. (e-mail: rituy@kth.se, nascetti@kth.se, yifang@kth.se) 15 September, 2025 }
}



\maketitle

\begin{abstract}
Earth Observation regression tasks such as building height, canopy height, and above-ground biomass estimation underpin critical applications in urban planning, forest monitoring, and climate policy, where both accuracy and reliability are critical. Yet most deep learning models yield only deterministic predictions, providing no indication of per-pixel reliability. These regression tasks are inherently challenging due to heterogeneous land surfaces, skewed target distributions, sensor noise, and signal saturation at high target values, making uncertainty (UC) estimation essential for reliable inference. We address this gap by modeling aleatoric uncertainty using year-long Sentinel-1 SAR and Sentinel-2 MSI time series, proposing two complementary approaches: (i) Gaussian UC, which jointly predicts mean and standard deviation under a Gaussian assumption, and (ii) Quantile UC, which estimates the 10th, 50th, and 90th quantiles to capture asymmetric and heteroscedastic error distributions. Both models are evaluated on three representative EO regression tasks at 10 m spatial resolution. 
Results show that both approaches match or surpass deterministic benchmarks and existing global products, while delivering well-calibrated, interpretable, and operationally useful confidence estimates. Notably, both models outperform the current 10 m state-of-the-art uncertainty-aware model for canopy height estimation. Our implementation will be available at: https://github.com/RituYadav92/EO-Regression-Uncertainty-Estimation

\end{abstract}

\begin{IEEEkeywords}
Uncertainty Estimation, Epistemic, Quantile, Gaussian, Time Series, Regression.
\end{IEEEkeywords}

\section{Introduction}
 \label{sec:intro}

\IEEEPARstart{U}{ncertainty} estimation has become central to trustworthy machine learning, particularly in safety-critical and large-scale deployment scenarios. In Earth Observation (EO), where satellite-derived products are used for decision-making in domains such as climate modeling, biodiversity monitoring, and urban planning, quantifying prediction uncertainty is equally essential. 
Errors arising from inherent sensor noise, atmospheric effects, or model limitations propagate through downstream applications; uncertainty estimates help distinguish reliable outputs from uncertain ones, improving trustworthiness and operational utility of EO products\cite{kendall2017uncertainties, paasche2022brave, gawlikowski2023survey}.

Despite significant progress in deep learning for EO, most models still produce single deterministic outputs, such as a class label for each pixel or a single scalar regression value per location. While aggregated evaluation metrics like RMSE or accuracy summarize performance over entire datasets, they do not convey the reliability of prediction at individual pixels. Uncertainty estimation addresses this gap by quantifying the variability or confidence associated with model outputs, enabling users to understand not only what a model predicts but also how certain it is about those predictions.

Predictive uncertainty formalizes this idea by characterizing the distribution over model outputs given an input, which may reflect either variability inherent in the data or uncertainty in the model itself. Depending on the source of variability, predictive uncertainty is typically decomposed into aleatoric and epistemic components. Aleatoric uncertainty stems from irreducible noise in the data, such as sensor errors, atmospheric interference, mixed pixels or inconsistencies in training labels. Epistemic uncertainty arises from limitations of the model itself, such as insufficient training data or overly restrictive parameterizations, and can in principle be reduced with more data or better models \cite{kendall2017uncertainties, hullermeier2021aleatoric, mobiny2021dropconnect}.In EO regression, strong irreducible aleatoric uncertainty is the dominant challenge due to atmospheric noise, mixed pixels, and sensor saturation, making explicit aleatoric modeling indispensable.

In computer vision, epistemic uncertainty is commonly captured via Bayesian neural networks \cite{springenberg2016bayesian, louizos2016structured, khan2018fast, teye2018bayesian}, Monte Carlo dropout as a Bayesian approximation \cite{gal2016dropout}, Laplace approximations \cite{ritter2018scalable, martens2015optimizing, botev2017practical}, test time data augmentation \cite{van2020uncertainty}, ensembles of deep models \cite{hansen2002neural, osband2016deep, lakshminarayanan2017simple, pearce2018high} and others. Aleatoric uncertainty is typically modeled by estimating a conditional probability distribution over the target variable given the inputs. A common choice is to parameterize this distribution as Gaussian, with the model jointly predicting its mean and variance \cite{lakshminarayanan2017simple}. More recent work has proposed alternative formulations, such as parameterizing prior distributions over predictive outputs \cite{malinin2018predictive} or learning auxiliary confidence scores \cite{corbiere2021confidence} and many others.
While these methods are well-established in computer vision, their systematic application to EO tasks has remained limited. Among EO studies, deep ensembles are the most widely adopted strategy, with reported applications in image classification \cite{lv2017remote}, scene classification \cite{dai2019semisupervised}, road segmentation \cite{haas2021uncertainty}, crop classification \cite{ahmed2023investigation, sonobe2018crop}, and canopy height estimation \cite{lang2022global}. Beyond ensembles, Gaussian likelihood models have been applied to canopy height estimation \cite{lang2022global} and biomass estimation \cite{weber2025unified}, while Bayesian neural networks have been employed for corn yield prediction \cite{ma2021corn} and image classification \cite{he2023bayesian}.
Despite the rapid growth of deep learning applications in EO, with hundreds of publications each year, investigations into predictive uncertainty tailored to EO tasks remain scarce.

Uncertainty estimation is particularly critical and more challenging for regression tasks. Unlike classification, where uncertainty is expressed as a probability distribution over discrete and finite classes, regression involves continuous-valued predictions with infinitely many possible outcomes in the target space, making it harder to define calibrated uncertainty measures \cite{tagasovska2019single, foong2019between}. Compared to classification, EO regression targets biomass density, building heights, and canopy height exhibit skewed or heavy-tailed distributions, complicating the modeling of predictive intervals \cite{white1978estimation, usui2021building, yadav2025high}. 
A Gaussian assumption may under-represent asymmetric uncertainty in such cases, whereas non-parametric methods such as quantile regression can better capture heteroscedasticity and skewness without distributional assumptions.

In this work, we explicitly model aleatoric uncertainty for EO regression using two complementary formulations:
\begin{enumerate}
    \item \textbf{Gaussian UC}: a parametric model that jointly predicts mean $\mu$ and standard deviation $\sigma$ to approximate the target distribution under a Gaussian assumption. This approach provides symmetric confidence intervals around mean. 
    This approach has been widely used in computer vision \cite{kendall2017uncertainties} and more recently adapted for EO tasks such as canopy height estimation \cite{lang2022global} and biomass mapping \cite{weber2025unified}. 
    \item \textbf{Quantile UC}, a non-parametric model that estimates the 10th, 50th, and 90th conditional quantiles ($Q10$, $Q50$, $Q90$) to capture potentially asymmetric and heteroscedastic prediction intervals. This enables models not only to predict the uncertainty of an estimate but also its potential direction.
\end{enumerate}
While the Gaussian method provides statistically well-defined and symmetric intervals, the quantile approach is better suited to the skewed and heavy-tailed distributions commonly found in EO regression tasks. 
The mean or median ($Q50$) captures the central tendency of the prediction, and the variance or spread ($Q10-Q90$) represents the uncertainty range around that estimate. Ideally, the mean should be close to the reference value, and the uncertainty should be small but realistic, such that prediction intervals cover the true value with the expected frequency. By comparing these two methods, we highlight their relative strengths in representing predictive uncertainty.

We evaluate these methods across three key EO regression tasks at 10m spatial resolution: building height estimation, canopy height estimation, and aboveground biomass estimation (AGB). For each task, a 12-month time series of Sentinel-1 SAR and Sentinel-2 MSI imagery (one image per month) is used as input.
Time series provide richer supervision than any single date imagery by capturing temporal variations in features of vegetation, buildings (shadow) and surroundings across different seasons. Importantly, time-series images collectively contain more information than any single image, due to differences in photometric or spatial coverage for instance \cite{okabayashi2024cross}. 

These three selected tasks are critical to applications ranging from urban planning, disaster risk assessment to biodiversity monitoring and carbon accounting. At the same time, they pose unique challenges due to heterogeneous morphology, seasonal variations, scale differences and signal saturation effects, among others. Prior work has addressed uncertainty for canopy height and biomass using uni-temporal GEDI LiDAR or single-date Sentinel data \cite{lang2022global,weber2025unified}; here, we fuse SAR and multispectral time series and extend uncertainty analysis to the urban domain of building height estimation. By benchmarking our Gaussian and quantile uncertainty models across these tasks, we provide a systematic assessment of uncertainty in EO regression tasks.

Our contributions are summarized as follows: 
\begin{enumerate}
\item We compare two deep learning regression models for uncertainty estimation based on Gaussian and quantile formulations of aleatoric uncertainty. 
\item We evaluate both approaches on three high-impact EO regression tasks covering natural and built environments: canopy height, above-ground biomass, and building height estimation.
\item We demonstrate two calibration metrics, "ErrorCoverage" and "DataCoverage", to evaluate the predicted uncertainty intervals from Gaussian and Quantile UC models, respectively.
\item We demonstrate that uncertainty estimation models achieve accuracy comparable to deterministic baselines while providing well-calibrated, interpretable, and operationally useful confidence estimates.
\end{enumerate}


\section{Study Areas}
\label{sec:Dataset}
Our study area covers five countries, namely Estonia, Finland, Germany, Netherlands and Switzerland. The reference labels are derived from the data provided by the official land and forest administrations of the country. The input data is year-long time series data from Sentinel-1 SAR and Sentinel-2 MSI. The dataset overview is given in Table \ref{data_spec}. The following subsections describe all the datasets utilized in this study.         

\vspace{\baselineskip}
\begin{table}[htbp]
\caption{Dataset Specifications. The target spatial resolution of all tasks is 10m and references are derived from airborne LiDAR or stereo orthophotos. BHE: Building Height Estimation; CHE: Canopy Height Estimation; BME: Biomass Estimation. }
\begin{center}
\resizebox{\columnwidth}{!}{%
\begin{tabular}{|c|c|c|c|}
\hline
&  &  &  \\[-1ex]
\textbf{Task} & \textbf{Dataset} & \textbf{Location} & \textbf{\# Train, Test Samples} \\ [0.75ex] \hline
&  &  &  \\[-1ex]
BHE & \cite{yadav2025high} & Netherlands, Estonia & 39000, 6000 \\ 
&  &  Switzerland, Germany &  \\[0.75ex] \hline
&  &  &  \\[-1ex]
BME & BioMassters\cite{nascetti2023biomassters} & Finland & 10227, 2773 \\ [0.75ex] \hline
&  &  &  \\[-1ex]
CHE & BioMassters ext. & Finland & 10227, 2773 \\ [0.75ex] \hline
\end{tabular}%
}
\label{data_spec}
\end{center}
\end{table}

\subsection{Building Height Estimation}
We used the dataset from \cite{yadav2025high}, containing $\approx$45000 patches over Estonia, Germany, Netherlands, and Switzerland. Each patch covers a $1280 \times 1280$ m\textsuperscript{2} area at 10 m spatial resolution. References are derived at 1 m resolution from airborne LiDAR and stereo orthophotos acquired nationwide over one to three years. The building heights have altimetric precision of $\pm$ 15 cm, $\pm$ 1 m, $\pm$ 50 cm, and $\pm$ 7 cm for Netherlands, Germany, Switzerland and Estonia, respectively.
The input data contains a 12-month time series of Sentinel-1 C band SAR and Sentinel-2 MSI data, where one image per month is provided. The Sentinel-1 SAR contains VV, VH channels from both ascending and descending orbits, whereas Sentinel-2 MSI are the least cloudy bottom-of-atmosphere images of each month, containing five bands (blue, green, red, near-infrared, and short-wave-infrared).
The dataset contains $\approx$ 39000 training patches and $\approx$ 6000 test patches. We used the 80/20 split for the training and validation from the provided training patches and test set is as provide.

\subsection{Above Ground Biomass Estimation}
We used BioMassters \cite{nascetti2023biomassters}, covering $\approx$13000 patches of Finnish forests. Each patch covers a 2560 $\times$ 2560 square meter area of forest and represents the yearly peak of AGB (the LiDAR surveys are measured during the summer months from July to August).
The reference above-ground biomass (AGB) data is based on Airborne LiDAR campaigns performed by the Finnish Forest Centre in cooperation with the National Land Survey (NLS) of Finland. Airborne LiDAR and aerial imagery are acquired following a six-year cycle to cover the entire country. AGB per cell is computed using calibrated allometric equations \cite{repola_biomass_2008} and expressed as tonnes per pixel at 10 m resolution.
The input data in BioMassters contains a 12-month time series of Sentinel-1 SAR and Sentinel-2 MSI. The Sentinel-1 images contain four bands, VV and VH bands from both ascending and descending orbits, whereas Sentinel-2 images are bottom-of-atmosphere images with all bands except bands 1, 9 and 10, as they contain atmospheric information irrelevant to AGB estimation.
We used the original train, validation and test splits provided by the BioMassters dataset.

\subsection{Canopy Height Estimation}
For the canopy height dataset, we collected canopy height measurements in Finnish forests at a 10 m spatial resolution for all AGB patches from the BioMassters dataset \cite{nascetti2023biomassters}. The canopy height data is derived from the same LiDAR source as used to derive above-ground forest biomass in the BioMassters dataset. In fact, canopy height is one of the attributes used in the allometric equation to derive AGB values. The dataset contains a total of $\approx$13000 patches, each patch covering a 2560 $\times$2560 square meter area of forest. We used Sentinel-1 and Sentinel-2 data from the BioMassters dataset as the input data for canopy height estimation task. The dataset splits are also kept the same.

\section{Methodology}
We propose two uncertainty regression models (1) Gaussian uncertainty, which predicts mean ($\mu$) and log variance ($\sigma^{2}$) of the target distribution, and (2) quantile regression, which estimates the 10th, 50th, and 90th quantiles ($Q10, Q50, Q90$), enabling asymmetric intervals that better capture skewed error distributions. Both models share a common deep learning base model but different uncertainty header modules. The subsection below explains the model architecture of the backbone and uncertainty header modules, followed by the training of the two models.

\subsection{Network Architecture}
\begin{figure*}[htbp]
    \centering
    \includegraphics[width=0.85\linewidth]{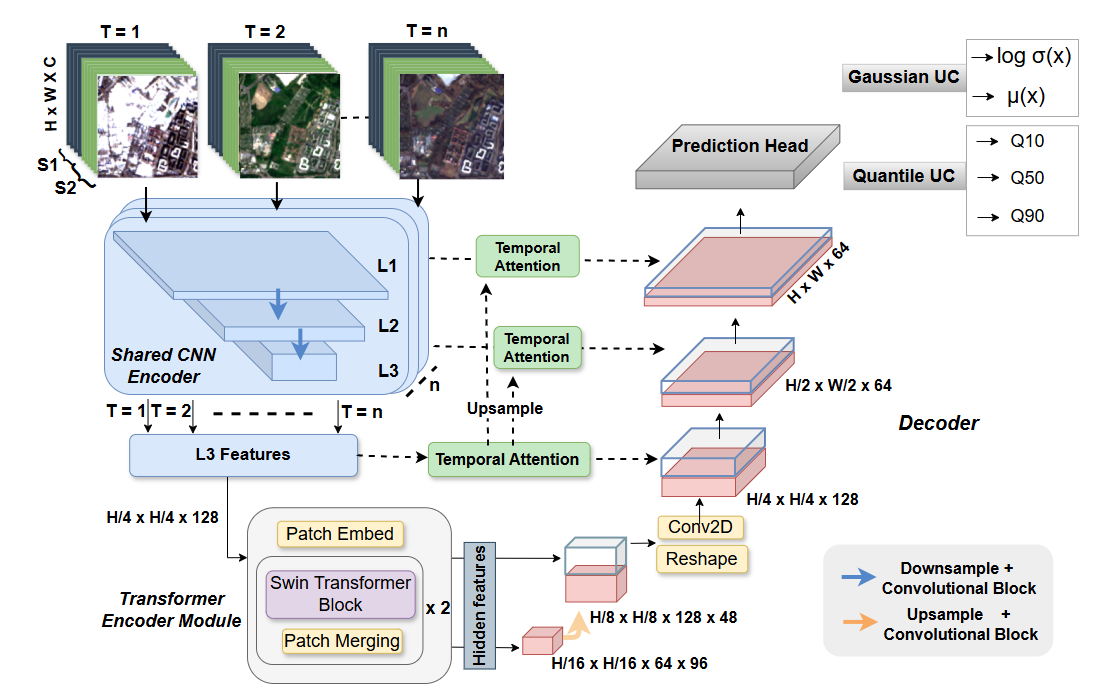}
    \caption{Network Architecture of Gaussian Uncertainty and Quantile Uncertainty Regression model. The base model is adapted from T-SwinUNet\cite{yadav2025high}.}
    \label{net_arch}
\end{figure*}

The network architecture of the base model is shown in Figure \ref{net_arch}. Its input $ x \varepsilon \mathbf{R}^{T \times C\times H \times W}$ is a time series of coregistered Sentinel-1 SAR and Sentinel-2 MSI images from the same geographical area. Here, $T$ corresponds to the temporal range of the time series, $C$ represents the spectral dimension and $H$, $W$ represent the height and width of the spatial dimension. We used $H=W=128$, corresponding to the patch size $128\times128$, $T=12$, corresponding to 12-month time series and $C=9$, corresponding to four Sentinel-1 SAR bands(S1) plus five Sentinel-2 MSI bands(S2). 
The network consists of a shared CNN encoder with ResNet50 backbone, one for each time stamp $T_{i}$. The output L3 features from the shared CNN encoder and applies a patch embedding layer to generate 3D tokens which are then mapped to the latent embedding space of size D = 48. The position embedding of the tokens is learnable here. With window size [7, 7, 7], patch size [2, 2, 2] and number of heads [3, 6], two consecutive Swin transformer blocks ~\cite{liu2021swin} are employed to implement multi-head attention with the shifted window technique. After each Swin transformer block, a patch merging layer is applied to downsample the output by a factor of 2. A patch merging layer concatenates the 2×2 neighboring patches and applies a linear layer on top. The output hidden features are upsampled, concatenated, reshaped and supplied to the decoder.

The decoder process features in three levels. At each level $(L1, L2, L3)$, the encoder feature maps are enhanced by temporal attention and concatenated with the same level feature maps from the decoder. After each convolutional block, a convolutional transpose layer is applied to upsample the features by a factor of two. On top of the output feature maps, a prediction head is attached; this head differs between the two uncertainty formulations (Gaussian UC and Quantile UC) and each is trained with a task-appropriate objective.

\subsection{Gaussian UC Regression Model and Training}
For the Gaussian UC model, the prediction head comprises two parallel convolutions with $3\times3$ kernel with padding = 1: one produces the pixel-wise log standard deviation map $\log \sigma$, and the other followed by a softplus produces pixel-wise mean map $\mu(x)$ with continuous values.

The model is trained under the assumption that the target $y$ at each pixel is drawn from a normal distribution with mean $\mu$ and variance $\sigma^2$. The primary loss used for training is per-pixel Negative Log-likelihood Loss (NLL) \cite{kendall2017uncertainties}. Our Gaussian UC model predicts a per-pixel mean map $\mu$ and a per-pixel log standard deviation $\eta = \log \sigma$. Now, the per-pixel negative log-likelihood is expressed in terms of $\eta$ as
\begin{equation}
\ell(y;\mu,\eta) \;=\; \eta + \tfrac{1}{2}\log 2\pi \;+\; \frac{(y-\mu)^2}{2\,\exp(2\eta)}.
\end{equation}

Let, $\mathcal{S}$ be all pixels with $|\mathcal{S}|=N$, and let $\mathcal{Z}=\{i\in\mathcal{S}\,:\,y_i\neq 0\}$ denote the subset of \textbf{nonzero} ground-truth pixels with $|\mathcal{Z}|=M$. To give more priority to nonzero pixels, we use the sum of mean NLL over all pixels $\mathcal{L}_{\text{all}}$ and again over all nonzero pixels $\mathcal{L}_{\text{nz}}$:
\begin{equation}
\mathcal{L}_{\text{all}} = \frac{1}{N} \sum_{i \in \mathcal{S}} \ell_i, 
\qquad 
\mathcal{L}_{\text{nz}} = \frac{1}{M} \sum_{i \in \mathcal{Z}} \ell_i,
\qquad 
\mathcal{L}_{\text{NLL}} = \mathcal{L}_{\text{all}} + \mathcal{L}_{\text{nz}}.
\end{equation}

To encourage sharp and correctly localized boundaries, we add an edge-preserving loss based on the Sobel operator \cite{paul2022edge}. For both the prediction $\mu$ and the reference map $y$, we compute horizontal and vertical gradient maps using the
standard Sobel filters, and penalize their differences: 
\begin{equation}
L_{\text{Sobel}} = \mathrm{mean}\left(
\left| S_x(\hat{y}) - S_x(y) \right|
+
\left| S_y(\hat{y}) - S_y(y) \right|
\right)
\end{equation}
where $S_x(\cdot)$ and $S_y(\cdot)$ denote the horizontal and vertical Sobel
gradient operators, respectively.

With low weight the RMSE loss is also used on the predicted mean $\mu$, given by:
\begin{equation}
\mathcal{L}_{\text{RMSE}} \;=\; \sqrt{\frac{1}{N}\sum_{i \in N}\bigl(\hat{y}_i - \mu_i\bigr)^2}\,
\end{equation}
The overall training objective for the Gaussian UC regression model is a weighted sum of NLL, RMSE, and Sobel edge losses:
\begin{equation}
\mathcal{L}_{\text{Gauss}}
= \alpha_1\,\mathcal{L}_{\text{NLL}}
+ \alpha_2\,\mathcal{L}_{\text{Sobel}}
+ \alpha_3\,\mathcal{L}_{\text{RMSE}},
\end{equation}
where $\alpha_1$, $\alpha_2$ and $\alpha_3$ are 2.0, 1.0 and 0.5 respectively. This weighting prioritizes the Gaussian NLL, uses RMSE as a regularizer, and leverages the Sobel term to preserve edges and suppress blur.

\subsection{Quantile UC Regression Model and Training}
For the Quantile UC regression model, the prediction head comprises three parallel convolutions with $3\times3$ kernel and padding = 1 that produce per-pixel three continuous quantile estimates $\hat q_{10,i}$, $\hat q_{50,i}$, and $\hat q_{90,i}$. For each pixel $i$ with ground truth $y_i$, the model predicts $\hat q_{10,i}$, $\hat q_{50,i}$, and $\hat q_{90,i}$. 

Training minimizes the asymmetric pinball (quantile) loss averaged over these three quantiles and all non-zero pixels. For quantile $q \in \{0.1,0.5,0.9\}$, prediction $\hat{Q}_{q,i}$, and error $u_i = y_i - \hat{Q}_{q,i}$, the pinball regression loss $ L_{\text{pin}}$ is defined by:
\[
\rho_q(u_i) =
\begin{cases}
q \, u_i, & u_i \geq 0, \\
(q - 1)\, u_i, & u_i < 0.
\end{cases}
\]

\[
\mathcal L_{\text{pin}} = \frac{1}{M} \sum_{i \in \mathcal{Z}} \sum_{q \in \{0.1,0.5,0.9\}} w_q \, \rho_q(y_i - \hat{Q}_{q,i}).
\]
Three regularizers added to the pinball loss are as follows:
\begin{enumerate}
\item Coverage shortfall: pushes the empirical coverage of $[\hat{Q}_{10},\hat{Q}_{90}]$ on non-zero pixels toward a nominal target $c^\star =0.8$. The coverage loss $L_{\text{cov}}$ is given below where, the indicator
function I(·) returns 1 if the condition holds and 0 otherwise
\[
\mathbf{I}_{i,\text{in}} = \mathbf{I} \big[ \hat{Q}_{10,i} \leq y_i \leq \hat{Q}_{90,i} \big], 
\hat c = \frac{1}{\mathcal{Z}}\sum_{i \in \mathcal{S}} \mathbf{1}_{i,\text{in}}.
\]

\[
\mathcal L_{\text{cov}} = (\hat c - c^\star)^2
\]
\item Non-crossing: enforces $\hat{Q}_{10}\le \hat{Q}_{50}\le \hat{Q}_{90}$ and the loss non-crossing loss $ L_{\text{nc}}$ is defined by:
\[
\mathcal L_{\text{nc}} = \frac{1}{S}\sum_{i \in N} 
\Big[ ReLU(\hat{Q}_{10,i}, \hat{Q}_{50,i}) + ReLU(\hat{Q}_{50,i}, \hat{Q}_{90,i}) \Big].
\]
\item Confidence-adaptive spread: shrinks the interval width $(\hat{Q}_{90}-\hat{Q}_{10})$ more when the median prediction is accurate, using an exponential weight. The spread loss $L_{\text{spread}}$ is given by :
\[
w_i \;=\; \exp\!\big(-\alpha\,\lvert \hat q_{50,i}-y_i \rvert\big),\qquad
\]
\[
L_{\text{spread}} \;=\; \frac{1}{N}\sum_{i \in N} \big(\hat q_{90,i}-\hat q_{10,i}\big)\, w_i,
\quad \alpha=10.
\]

\end{enumerate}
\noindent The total loss $\mathcal{L}_{\text{pinTotal}}$ combines the pinball loss with three regularization terms, as defined below. The weighting coefficients $\lambda_{\text{cov}}, \lambda_{\text{nc}}, \lambda_{\text{spread}}$ are learnable and initialized to $0.1$, $0.2$, and $0.3$, respectively.
\begin{equation}
L_{\text{pinTotal}}
\;=\; \,L_{\text{pin}}
\;+\; \lambda_{\text{coverage}}\,L_{\text{cov}}
\;+\; \lambda_{\text{noncross}}\,L_{\text{nc}}
\;+\; \lambda_{\text{spread}}\,L_{\text{spread}},
\end{equation}

\noindent To stabilize the median estimate and preserve edges, we apply the RMSE loss $\mathcal{L}_{\text{RMSE}}$ and the Sobel edge loss $\mathcal{L}_{\text{Sobel}}$ to the median prediction $Q50$. The overall training objective for the quantile UC regression model is a weighted sum of the pinball loss and these two terms, defined as follows, where $\beta_1$, $\beta_2$ and $\beta_3$ are 2.0, 1.0 and 0.5, respectively.

\begin{equation}
\mathcal{L}_{\text{Quant}}
= \beta_1\,\mathcal{L}_{\text{pinTotal}}
+ \beta_2\,\mathcal{L}_{\text{Sobel}}
+ \beta_3\,\mathcal{L}_{\text{RMSE}},
\end{equation}

\subsection{Training and Implementation Details}
\label{subsec:imp_details}
The time series input was augmented by adding a random channel drop (noise) with 0.2 probability. The added noise has a regularization effect during training, which in turn helps to reduce overfitting. 
All hyperparameters were fine-tuned on the training and validation datasets, and the evaluation was done on the held-out test sets. The Gaussian UC model has a total of 2818272 trainable parameters and quantile UC model has 3578081 trainable parameters. 
Models were trained for 100 epochs with a batch size of 4. We used AdamW optimizer with an initial learning rate of 0.0001, a minimum learning rate of 0.000001, a decay rate of 0.5 and an early stop patience of 10 epochs. The learning rate decay and early stopping was controlled by the "reduce on plateau" method. For regularized training, we used layer dropout with a rate of 0.1. The code was implemented in PyTorch. The experiments were carried out on an NVIDIA GeForce RTX 3080 GPU.

\subsection{Evaluation Metrics}
\label{sec:eval}
In the Gaussian UC model, pixel-wise predicted mean values ($\mu$) are evaluated using root mean square error (RMSE), $R^{2}$ score, and mean absolute percentage error (MAPE) and intersection over union (IoU). The first four metrics are used to evaluate predicted mean ($\mu$) building height, canopy height and biomass estimation values against labels $gt$. 
We also use the Normalized Median Absolute Deviation (nMAD), a robust evaluation metric commonly used in regression tasks, especially when data contains outliers and can inflate RMSE with a few bad outliers, e.g., incorrect estimation of high-rise buildings or wrong ground. It captures the scatter of errors (r) around the median, formulated as
\begin{equation}
\mathrm{NMAD} = 1.4826 \times \mathrm{median}\left( \, | r_i - \mathrm{median}(r) | \, \right),
\quad r_i = gt_i - \mu_i
\end{equation}
Low nMAD indicates height estimates are consistently close to reference heights across most buildings. Also, the gap between RMSE and nMAD reflects the presence of outliers or a symmetric distribution.

The predicted standard deviation $\sigma$ values are evaluated against the error between predicted mean values $\mu$ and the labels $gt$ under Gaussian assumption. The evaluation is done to estimate the percentage error covered under one, two and three $\sigma$ values. The Error Coverage under $k\sigma$ $(k = 1,2,3)$ is defined as
\begin{equation}
\text{ErrorCoverage}(k\sigma) =
\frac{1}{N} \sum_{i=1}^{N}
\mathbf{I}\!\left( |gt_i - \mu_i| \leq k \cdot \sigma_i \right) \times 100,
\end{equation}
where the dataset reference label contains $N$ nonzero pixels, each sample $i$ with reference label $gt_i$, predicted mean $\mu_i$, and predicted standard deviation $\sigma_i$. The indicator function $\mathbf{I}(\cdot)$ returns $1$ if the condition holds and $0$ otherwise.

In the quantile model, the median prediction ($Q50$) is evaluated using RMSE, $R^{2}$, MAPE, and nMAD, similar to the evaluation used for the Gaussian UC model. 
The predicted uncertainty interval [$Q10$, $Q90$] is evaluated against labels $gt$ as follows:
\begin{equation}
\text{DataCoverage}(Q10, Q90) = 
\frac{1}{N} \sum_{i=1}^N 
\mathbf{I}\!\left( Q10_i \leq gt_i \leq Q90_i \right) \times 100.
\end{equation}

A well-calibrated model should predict mean $\mu$ (or median $Q50$) closer to the target labels $gt$, with residuals falling within the predicted intervals at the expected rates. Specifically, the ErrorCoverage in $\pm1\sigma$, $\pm2\sigma$, and $\pm3\sigma$ should approach the nominal Gaussian coverage of $68.27\%$, $95.46\%$ and $99.73\%$ respectively, while the DataCoverage of the quantile interval [$Q10$–$Q90$] should approach the nominal $80\%$. The width of the quantile interval or $\pm \sigma$ reflects the uncertainty in predicted mean or median value, i.e., a larger interval indicates higher uncertainty and a narrower interval indicates lower uncertainty or higher confidence in the prediction.

\section{Results}
\label{sec:results}
We evaluate two uncertainty estimation models: (i) Gaussian UC, predicting mean and standard deviation under a Gaussian assumption, and (ii) Quantile UC, predicting $Q10$, $Q50$ and $Q90$ without a distributional assumption. These two uncertainty estimation approaches are tested on three remote sensing tasks: i) building height estimation, ii) canopy height estimation, and iii) above-ground biomass estimation. 
The summary of the overall results is provided in Table \ref{Compiled_results}. 

\begin{table*}[t]
  \caption{Summarized test results on all three tasks, Building Height Estimation (BHE), Canopy Height Estimation (CHE) and BioMass Estimation (BME). RMSE and nMAD for building and canopy height estimations are in meters per pixel, whereas for biomass estimation, the unit is tons per pixel. These results are on the held-out test sets.}
  \label{Compiled_results}
  \centering
  \resizebox{0.75\textwidth}{!}{%
\begin{tabular}{ccccccccc}
\hline
\multicolumn{1}{l}{\multirow{2}{*}{ Gaussian UC}}  & \multirow{2}{*}{\textbf{RMSE $\downarrow$}} & \multirow{2}{*}{\textbf{nMAD} $\downarrow$} & \multirow{2}{*}{\textbf{$R^{2}$} $\uparrow$} & \multirow{2}{*}{\textbf{MAPE} $\downarrow$} & \multirow{2}{*}{\textbf{IoU} $\uparrow$} & \multicolumn{3}{|c}{\textbf{ErrorCoverage $\pm$ std.}} \\ \cline{7-9}
\multicolumn{1}{l}{} &  &  &  &  &  &  \multicolumn{1}{|c}{\textbf{$\pm1$ std.}}  & \textbf{$\pm2$ std.} & \textbf{$\pm3$ std.}        \\ \hline
&  &  &  &  &  &  &  &\\[-2ex] 
\textbf{BHE }   & 2.56   & 2.34 & 0.42   & 35.41   & 0.52   & \multicolumn{1}{|c}{0.68}   & 0.94   & 0.99    \\
&  &  &  &  &  &  \multicolumn{1}{|c}{} &  &\\[-2ex]
\textbf{CHE }  & 2.33   & 1.49 & 0.90   & 17.00   & 0.94   &  \multicolumn{1}{|c}{0.82}  & 0.94   & 0.96     \\
&  &  &  &  &  &  \multicolumn{1}{|c}{} &  &\\[-2ex]
\textbf{BME }    & 30.14   & 20.25 & 0.80   & 33.01  & 0.99   &  \multicolumn{1}{|c}{0.75}  & 0.96  & 0.99     \\
&  &  &  &  &  &  \multicolumn{1}{|c}{} &  &\\[-2ex]
\hline
&  &  &  &  &  &  &  &\\[-2ex]
Quantile UC & \textbf{RMSE $\downarrow$} & \textbf{nMAD} $\downarrow$ & \textbf{$R^{2}$} $\uparrow$ & \textbf{MAPE} $\downarrow$ & {\textbf{IoU} $\uparrow$} & \multicolumn{3}{|c}{\textbf{DataCoverage [$Q10$, $Q90$] }}        \\ 
&  &  &  &  &  &  &  &\\[-2ex]  \hline 
&  &  &  &  &  &  &  &\\[-2ex] 
\textbf{BHE }  & 2.37   & 1.93 & 0.50   & 33.17   & 0.62   & \multicolumn{3}{|c}{0.82}  \\
&  &  &  &  &  &  \multicolumn{1}{|c}{} &  &\\[-2ex]
\textbf{CHE }  & 2.54   & 2.00 & 0.87   & 19.10   & 0.94   & \multicolumn{3}{|c}{0.79}  \\
&  &  &  &  &   &\multicolumn{1}{|c}{} &  &\\[-2ex]
\textbf{BME }    & 32.66   & 21.00 & 0.76   & 36.20  & 0.99   & \multicolumn{3}{|c}{0.82}  \\
&  &  &  &  &  &  &  &\\[-2ex]  \hline 
  \end{tabular}%
 }
\end{table*}

\begin{figure*}[!t]
    \centering
    \includegraphics[width=0.90\linewidth]{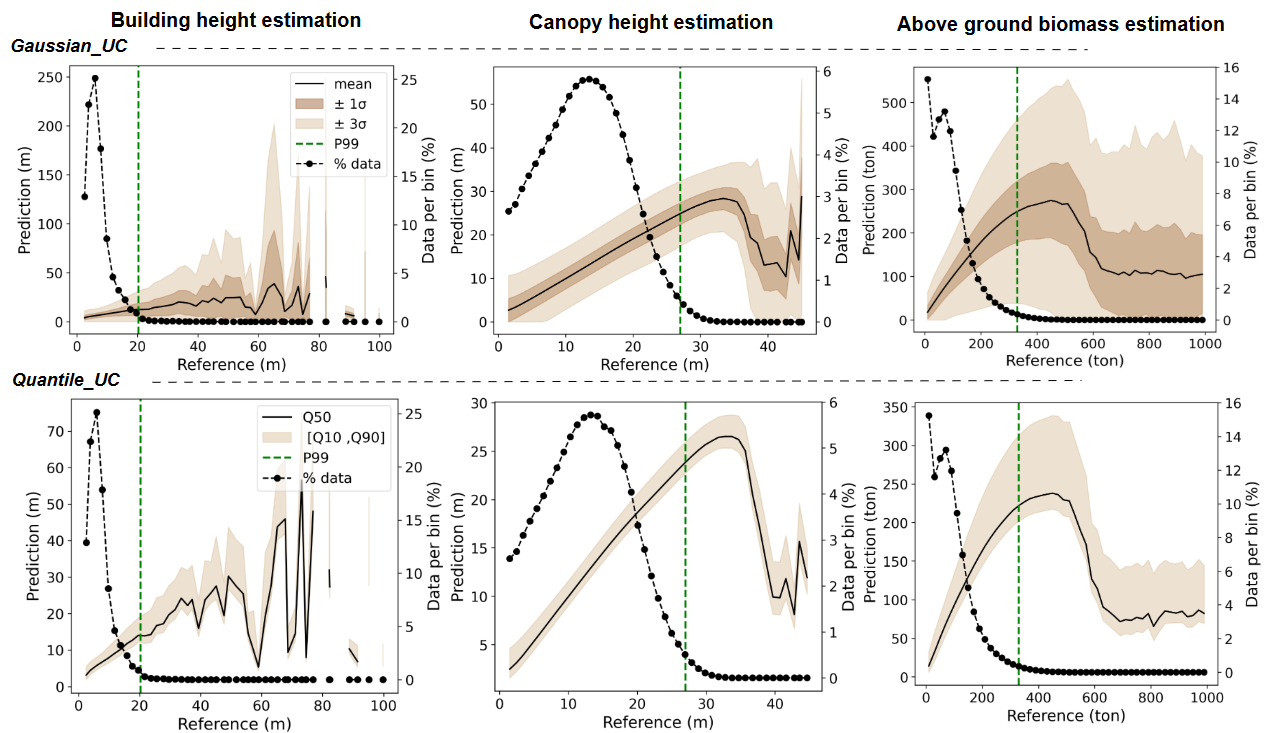}
    \caption{Predictions with Uncertainty and Data Distribution. The plots visualize the trends in predictions and uncertainty relative to the distribution of test data. P99 refers to the 99 percentile of the data.}
    \label{fig:coverage_plots}
\end{figure*}

\begin{figure*}[!t]
    \centering
     \includegraphics[width=0.93\linewidth]{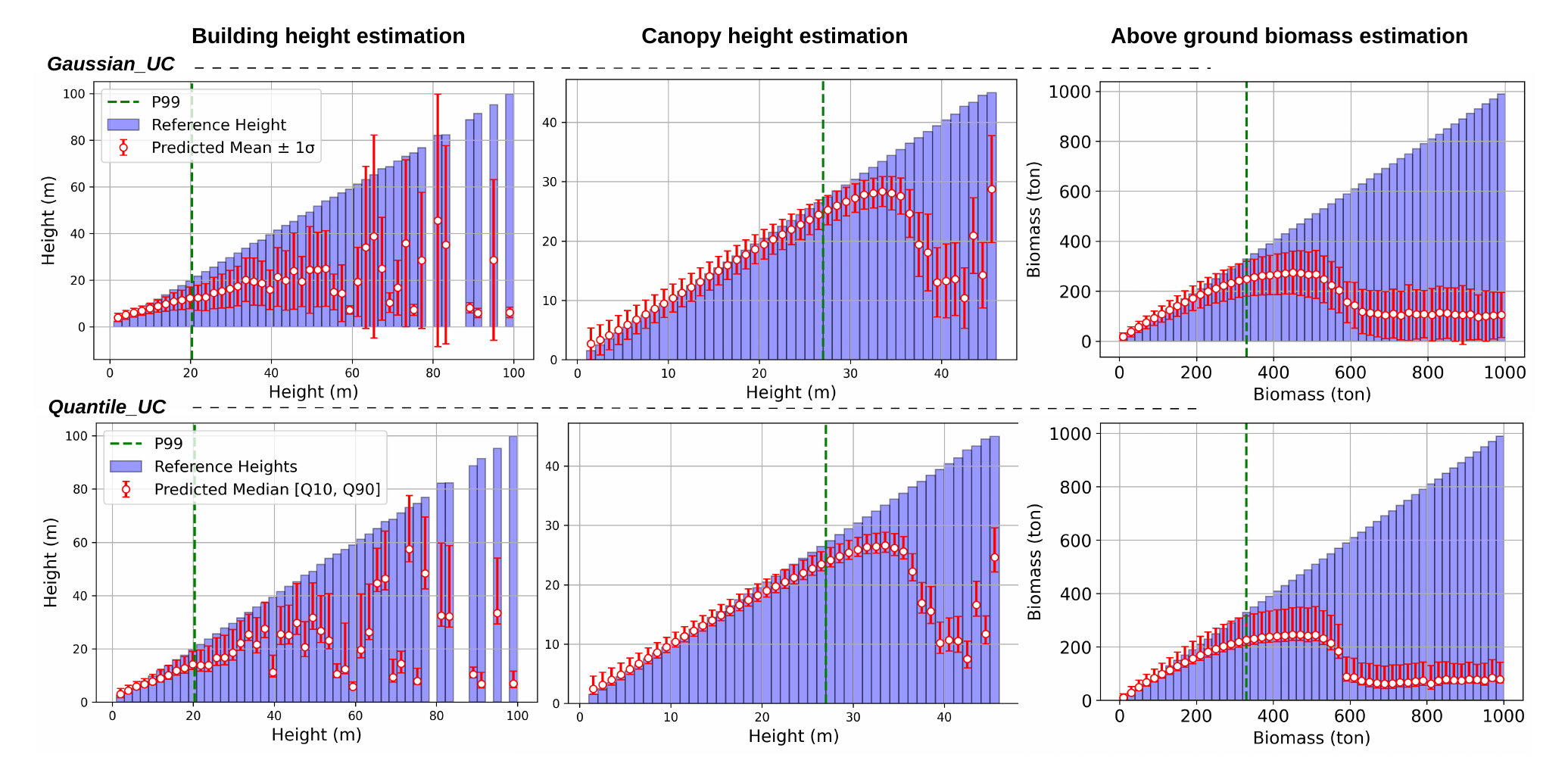}
    \caption{Bin-wise plot of Predicted vs. Reference Values with Uncertainty range. Each bin is 2m wide.}
    \label{fig:binplot}
\end{figure*}

Overall, building and canopy heights are predicted with an RMSE of $\approx$2.5 m per pixel, while biomass has a higher error of $\approx$35 tons per pixel. The gap between RMSE and nMAD scores indicates that biomass estimation is more prone to outliers than building and tree canopy height estimation.
Canopy height and biomass achieve higher $R^{2}$ scores (0.8–0.9), suggesting that the model learned variation in vegetation height efficiently, while learning complex building height variations remains more challenging ($R^{2}\approx$0.5). This is consistent with bin-wise plots in Figures \ref{fig:coverage_plots} and \ref{fig:binplot}, where canopy and biomass estimates align more closely with the diagonal. 
Across metrics, the Gaussian UC model performs best for canopy height and biomass, while Quantile UC gives slightly better scores for building height (lower RMSE and nMAD, higher $R^{2}$). This likely reflects the heterogeneity of urban environments, where localized biases and non-Gaussian or skewed distributions are better captured by asymmetric prediction intervals, inherent in the quantile UC model \cite{angelopoulos2022image}, whereas the Gaussian assumption is more appropriate for the smoother vegetation-related targets.

Figure \ref{fig:coverage_plots} and Figure \ref{fig:binplot} show that for building height and biomass, the upper part of the interval ($Q50$–$Q90$) is typically wider than the lower part ($Q10$–$Q50$), which aligns with the right‑skewed error distributions and systematic underestimation at high target values. This directional behavior allows the model to express that large positive errors are more likely than large negative ones, something a symmetric Gaussian interval cannot capture.

The proposed UC models not only estimate the target regression values but also quantify how uncertain or noisy those predictions are, a critical skill of probabilistic regression. 
The Gaussian UC model quantifies prediction uncertainty via the predicted standard deviation $\sigma$, evaluated against the residuals between predicted mean and reference values through the ErrorCoverage score. Across the three tasks, the ErrorCoverage score shows that 68–82\% of residuals/errors fall within the first predicted standard deviation, 94–96\% within the second, and 96–99\% within the third, which is close to the nominal Gaussian coverage. In some cases, however, the $3\sigma$ intervals extend beyond plausible values (Figure \ref{fig:coverage_plots}), indicating that the model occasionally compensates for tail misfit by inflating uncertainty in sparsely sampled high-value regions.
For the Quantile UC model, uncertainty is measured via DataCoverage over the [Q10–Q90] interval. The quantified DataCoverage scores are 0.82, 0.79, and 0.82 for BHE, CHE, and BME tasks, respectively. These scores are computed with a small tolerance around the predicted intervals (1 m for building and canopy height, 10 tons for biomass, more results in Table 1, Supplementary material). The scores are close to the nominal 80\% expected for the 10th–90th quantile range, confirming that the asymmetric quantile intervals are both directionally informative and well calibrated. By design, however, the [Q10–Q90] interval accounts for 80\% of the data, leaving the remaining 20\% outside the predicted range.

Figure \ref{fig:coverage_plots} and Figure \ref{fig:binplot} provide a more detailed view of how prediction errors and uncertainty behave across the range of reference values. For all tasks, underestimation of the mean or median increases toward higher target values, and the associated uncertainty intervals (standard deviations for Gaussian UC, [$Q10$, $Q90$] for Quantile UC) also widen, partially capturing these larger residuals. Beyond the saturation point (very tall buildings, very high canopy or biomass), both models tend to underestimate strongly and the intervals become wider but still fail to cover all large errors, which explains the residual miscalibration in the extreme tail. 
At the low end of the range, the MAPE plots (Figure 2, Supplementary material) show elevated relative errors, and together with Figure \ref{fig:binplot} they indicate overestimation in the low range. The overall MAPE trends show that both models have difficulties in predicting values in the early low-range and high-range target values. This might be attributed to feature confusion in the low range (close to 0) and to data deficiency coupled with saturation effects in the high range. 

For building height estimation task, the Quantile model gives sharper and more adaptive estimates, particularly around edges and tall structures, though some tall buildings remain outside the predicted range.
In contrast, the Gaussian UC model produces smoother predictions with wider uncertainty intervals, especially in the upper height ranges, which reflects the difficulty of extrapolating to rare, high-rise buildings.
For canopy height estimation, the quantile model provides overall better results up to 37m height, with similar median or mean predictions but smaller and directional uncertainty intervals accurately capturing the errors with high confidence. After 37m, both models struggle to estimate heights accurately; however, the Gaussian model performs slightly better, also supported by larger uncertainty intervals.
For biomass estimation, the Gaussian model provides more accurate predictions and uncertainty intervals that scale with error. Similar to canopy height, both models struggle beyond a certain point ($\approx$ 500 tons). The predictions saturate and uncertainty intervals widen without fully covering the large errors. 

The data distribution curve in Figure \ref{fig:coverage_plots} further shows that the 99th percentile of the data lies in the lower-to-mid range, where both models achieve accurate predictions with well-calibrated uncertainty, while the largest prediction discrepancies appear in the high-value tail, where the estimated uncertainties are least reliable.
Overall, the Quantile UC model is particularly advantageous for building height estimation and for capturing asymmetric, right‑skewed error patterns, while the Gaussian UC model tends to achieve better accuracy and more stable coverage for canopy height and biomass.

\subsection{Qualitative Analysis}
\begin{figure*}[!t]
    \centering
    \includegraphics[width=1.0\linewidth]{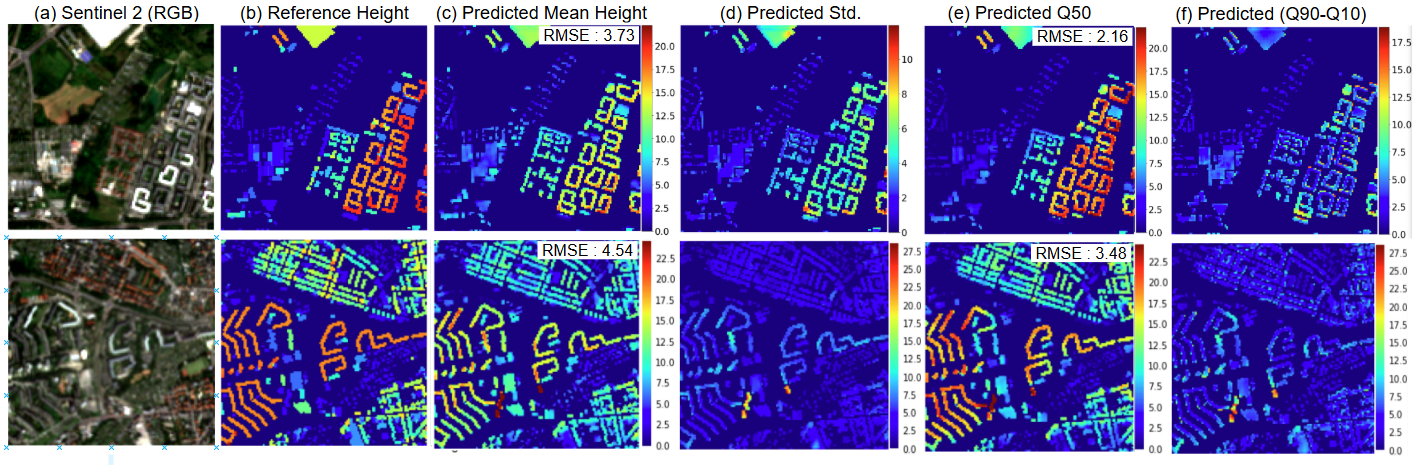}
    \caption{Gaussian UC vs. Quantile UC sample results on Building Height estimation. Gaussian UC: (c) predicted mean, (d) std deviation. Quantile UC: (e) predicted Q50, (f) Q90-Q10 interval. Higher std or Q90-Q10 reflects higher uncertainty. }
    \label{fig:vis_BHE}
\end{figure*}

\begin{figure*}[!t]
    \centering
    \includegraphics[width=1.0\linewidth]{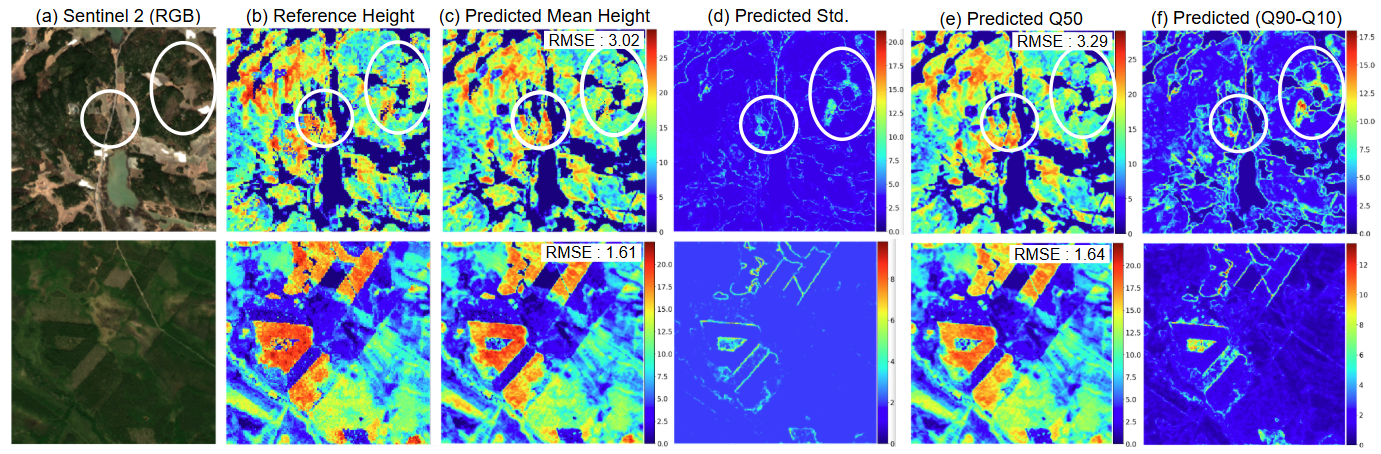}
    \caption{Gaussian UC vs. Quantile UC sample results on Canopy Height Estimation}
    \label{fig:vis_CHE}
\end{figure*}

\begin{figure*}[!t]
    \centering
    \includegraphics[width=1.0\linewidth]{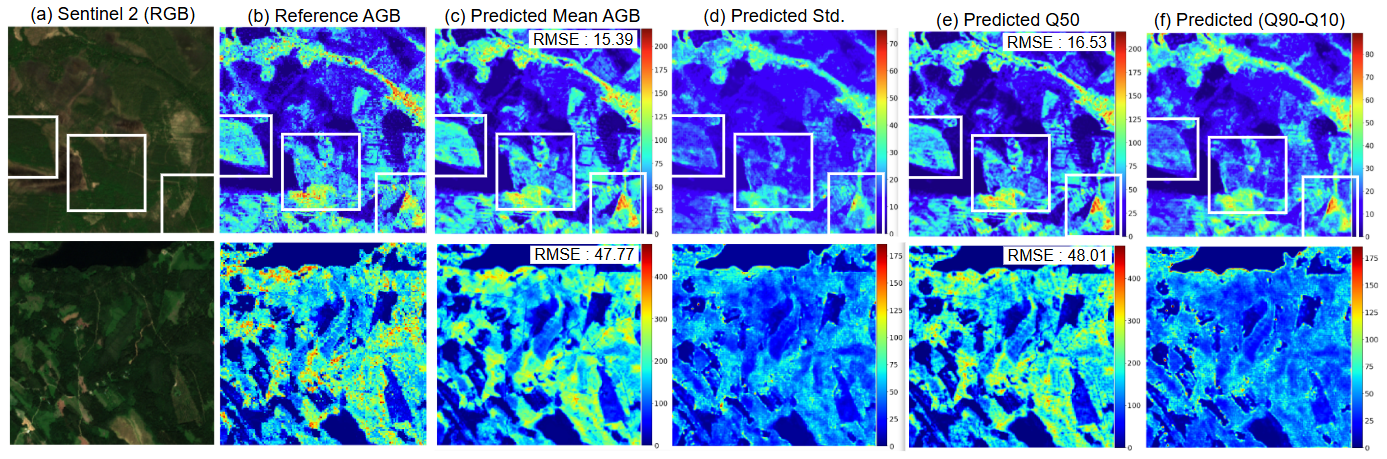}
    \caption{Gaussian UC vs. Quantile UC sample results on Above-ground Biomass Estimation}
    \label{fig:vis_BME}
\end{figure*}

For Qualitative analysis, three samples are visualized in Figure \ref{fig:vis_BHE}, \ref{fig:vis_CHE} and \ref{fig:vis_BME}. These samples illustrate that both the predicted mean from the Gaussian UC model and the predicted median ($Q50$) from the quantile UC model broadly capture the spatial patterns present in the reference maps, but their relative performance differs. 

For building height estimation (Figure~\ref{fig:vis_BHE}), the Quantile UC model produces sharper and more spatially detailed predictions, accurately distinguishing low, mid, and high-rise buildings while preserving height variations and structural boundaries. 
In contrast, the Gaussian UC model tends to underestimate absolute building heights, though the relative height differences between buildings are generally preserved. Regarding uncertainty, the Quantile interval $(Q_{90} - Q_{10})$ correlates well with prediction errors: it is wide in regions of high discrepancy, particularly along building edges where boundary pixels are typically underestimated, and narrow where predictions are accurate.
This behavior reflects a coherent uncertainty estimate, as the model signals high confidence when its prediction is reliable and expresses higher uncertainty when it is not. For the Gaussian UC model, the $1\sigma$ intervals are narrow but still cover most errors within one standard deviation; the remaining errors fall within the lower-confidence $2\sigma$ and $3\sigma$ ranges.

For canopy height estimation (Figure \ref{fig:vis_CHE}), both models provided similar height estimates with few underestimation areas (highlighted), which are well captured by the corresponding uncertainty intervals.
Similar to building height estimation tasks, quantile intervals are also larger on canopy boundaries. These large uncertainty intervals in high-contrast areas, such as boundaries, reflect the intrinsic uncertainty in localizing edges. This is probably due to the low spatial resolution of input features. 
Although this effect should occur in both models, the Gaussian approach tends to keep $\sigma$ relatively low to avoid loss inflation, resulting in narrower variance estimates but smoother and wider boundaries.

Above-ground biomass estimation (Figure \ref{fig:vis_BME}). Both models broadly capture the spatial distribution of biomass, but the Quantile UC model better preserves fine-grained gradients from low to high biomass values, with more evident local over- and underestimation (e.g., overestimation in the lower-right, underestimation in the centre of the sample). The Gaussian UC model, in contrast, produces smoother predictions that reflect the averaging nature of mean estimation. In terms of uncertainty, both models show well-correlated intervals: the predicted standard deviation and quantile intervals widen for high-error pixels and narrow for low-error pixels, confirming that the uncertainty estimates are spatially meaningful across the biomass range.

\begin{table*}[t]
  \caption{Comparison with existing methods at 10m spatial resolution. Evaluation is done on held-out test sets, and the best scores are highlighted.}
  \label{SOTA_comp}
  \centering
  \resizebox{0.77\textwidth}{!}{%
\begin{tabular}{ccccccccc}
\hline
\multicolumn{1}{l}{\multirow{2}{*}{ }}  & \multirow{2}{*}{\textbf{RMSE $\downarrow$}} & \multirow{2}{*}{\textbf{nMAD} $\downarrow$} & \multirow{2}{*}{\textbf{$R^{2}$} $\uparrow$} & \multirow{2}{*}{\textbf{MAPE} $\downarrow$} & \multirow{2}{*}{\textbf{IoU} $\uparrow$}& \multicolumn{2}{|c}{\textbf{ErrorCoverage}} &\textbf{DataCoverage} \\ 
\multicolumn{1}{l}{} &  &  &  &  &  &  \multicolumn{2}{|c}{\textbf{$\pm1, 2, 3$ std.}}  & [Q10, Q90]        \\ \hline
&  &  &  &  &  &  &&  \\[-2ex] 

\multicolumn{9}{c}{Building Height Estimation} \\
&  &  &  &  &   &  &  &\\[-2ex]   \hline
&  &  &  &  &  &  &  &\\[-2ex]
Gaussian UC   & 2.56   & 2.34 & 0.42   & 35.41   &  0.52 & \multicolumn{2}{|c}{0.68, 0.94, 0.99} &  -  \\
&  &  &  &  &  &  \multicolumn{1}{|c}{} &  &\\[-2ex]
Quantile UC)   & \textbf{2.37}   & \textbf{1.93} & 0.50  & \textbf{33.17}   &  \textbf{0.62}  &  \multicolumn{2}{|c}{-} &  0.81          \\
&  &  &  &  &  &  \multicolumn{2}{|c}{} &  \\[-2ex]
T-SwinUNet   & \textbf{2.35}   & \textbf{1.93} & \textbf{0.52}   & \textbf{32.10}   &  \textbf{0.62} & \multicolumn{2}{|c}{-}  &  -  \\
&  &  &  &  &  &  \multicolumn{2}{c}{} &  \\[-2ex]  \hline
&  &  &  &  &  &  &  &\\[-2ex]

\multicolumn{9}{c}{Canopy Height Estimation} \\
&  &  &  &  &  &   &  &\\[-2ex]   \hline
&  &  &  &  &  &  &  &\\[-2ex]
&  &  &  &  &  &  \multicolumn{1}{|c}{} &  &\\[-2ex]
Gaussian UC   & \textbf{2.33}   & \textbf{1.49} & \textbf{0.90}   & \textbf{17.00}   &  \textbf{0.94} & \multicolumn{2}{|c}{0.82, 0.94, 0.99} &  -  \\
&  &  &  &  &  &  \multicolumn{2}{|c}{} &  \\[-2ex]
Quantile UC   & 2.54   & 2.00 & 0.87   & 19.10   &  \textbf{0.94}  & \multicolumn{2}{|c}{-} &  0.79      \\
&  &  &  &  &  &  \multicolumn{2}{|c}{} &  \\[-2ex]
Lang et al.\cite{lang2023high}   & 5.54   & 5.34 &  0.13   &   41.28   &  0.81 & \multicolumn{2}{|c}{0.79, 0.94, 0.96} &  -  \\
&  &  &  &  &  &  \multicolumn{1}{|c}{} &  &\\[-2ex]
tolan et al.\cite{tolan2024very}   & 7.96   & 8.72 &  -0.8   &   100.00   &  0.77 & \multicolumn{2}{|c}{-} &  -   \\
&  &  &  &  &  &  \multicolumn{2}{c}{} &  \\[-2ex]  \hline
&  &  &  &  &  &  &  &\\[-2ex] 

\multicolumn{9}{c}{Above ground Biomass Estimation} \\
&  &  &  &  &  &  &\\[-2ex]   \hline
&  &  &  &  &  &  &\\[-2ex]
Gaussian UC   & \textbf{30.14}   & \textbf{20.26} & \textbf{0.82}   & 33.15  &  0.72 &  \multicolumn{2}{|c}{0.75, 0.96, 0.99} &  -  \\
&  &  &  &  &  &  \multicolumn{2}{|c}{} &  \\[-2ex]
Quantile UC   & 32.66   & 21.00 & 0.76   & 36.20   &  \textbf{0.92} & \multicolumn{2}{|c}{-} &  0.82     \\
&  &  &  &  &  &  \multicolumn{2}{|c}{} &  \\[-2ex]
Biomassters    & 31.08   & 28.47 & 0.80   & \textbf{31.17}   &  0.78 & \multicolumn{2}{|c}{-}  & -  \\
&  &  &  &  &  &  \multicolumn{2}{c}{} &  \\[-2ex]  \hline 
  \end{tabular}%
}\end{table*}
\subsection{Comparison with competing Methods and products}
\label{sec:SOTA}
Our Gaussian UC and Quantile UC results are compared against recent state-of-the-art methods at 10 m resolution, prioritizing models with uncertainty estimation that have open-source code or pre-reported results on our test sets.
For building height, no uncertainty-estimation baseline exists, so we compare against the deterministic T-SwinUNet \cite{yadav2025high} benchmarking model on our used dataset. For canopy height, we compare against an existing uncertainty-based canopy height estimation model \cite{lang2023high}, which predicts pixel-wise ($\mu$, $\sigma$) from Sentinel-2 inputs using a Gaussian likelihood head on a U-Net backbone with a deep ensemble of five models, analogous to our Gaussian UC approach. In addition, the results are also compared with the recent 1 m canopy height maps from Meta \cite{tolan2024very}, produced by a Vision Transformer with a convolutional decoder trained on very-high-resolution (50 cm) RGB imagery. For above-ground biomass, we evaluate our results against deterministic benchmarks on the BioMassters dataset \cite{nascetti2023biomassters}.  Results are shown in Table \ref{SOTA_comp}.

Both Gaussian and Quantile UC models match or surpass deterministic baselines while providing calibrated, pixel-wise confidence. For building height, the Quantile UC model performs on par with the benchmarking model (T\text{-}SwinUNet), whereas the Gaussian UC model is slightly weaker on all metrics (RMSE, nMAD, R2 and MAPE). 
Gaussian UC model provides reliable coverage within the $1\sigma$ interval, but its higher confidence intervals ($2\sigma$, $3\sigma$) are often excessively wide, especially after 99 percentile height.
For canopy height, both Gaussian and Quantile UC models outperform the two global reference products at 10 m and 1 m spatial resolutions. The estimates from \cite{lang2023high} yield consistently low scores across all metrics, though the associated $\sigma$ values produce error coverage comparable to our Gaussian UC model. Although the recent 1 m global canopy height product by \cite{tolan2024very} provides accurate canopy masks, its height estimates perform poorly on the Finland test set across all metrics.
On above-ground biomass, both Gaussian UC and Quantile UC are competitive relative to the BioMassters top-performing model, and Gaussian UC achieves the strongest overall performance. Although BioMassters reports a lower MAPE (31.17), its nMAD is markedly higher (28.47) than our Gaussian UC and Quantile UC models, indicating that our uncertainty models are more robust than the deterministic top-performing model on BioMassters.

\section{Conclusion}
\label{sec:conclusion}
Our comparative analysis of Gaussian and Quantile uncertainty models shows that uncertainty estimation can improve the reliability and interpretability of EO regression products. Gaussian UC provides accurate pixel-level estimates and well-calibrated uncertainty intervals for vegetation-related tasks, while Quantile UC better captures the heterogeneity of urban environments with local biases and skewed distributions. An additional advantage of the Quantile model is that it provides the direction of uncertainty (upper or lower quantile interval), which can be valuable in applications where overestimation or underestimation has different consequences.
Across three tasks, the width of the predicted uncertainty intervals is well aligned with the distribution of errors, demonstrating their utility beyond pixel-level regression and confirming their value for large-scale monitoring. Both approaches reveal trade-offs between coverage and sharpness of uncertainty bounds, suggesting that future work should focus on hybrid strategies that combine their strengths.
By benchmarking these methods across diverse EO applications, we demonstrate the importance of explicitly modeling aleatoric uncertainty to increase the trustworthiness and usability of EO-derived products. Beyond improving predictive reliability, these approaches support more informed decision-making in urban planning, ecosystem monitoring, and climate policy. Future work will extend this framework by integrating the two methods, testing robustness under distribution shifts, and scaling to global applications.

\bibliographystyle{IEEEtran}
\bibliography{refer}
\end{document}